\documentclass[lettersize,journal]{IEEEtran}
\usepackage{amsmath,amsfonts}
\usepackage{algorithmic}
\usepackage{algorithm}
\usepackage{graphicx}
\usepackage{subfigure}
\usepackage{array}
\usepackage{subcaption}
\usepackage{textcomp}
\usepackage{url}
\usepackage{amssymb}
\usepackage{verbatim}
\usepackage{hyperref}
\usepackage{balance}
\usepackage{graphicx}
\usepackage{cite}
\usepackage{xcolor}
\usepackage{multirow}
\newtheorem{theorem}{Theorem}
\newtheorem{lemma}{Lemma}
\newtheorem{definition}{Definition}

\newtheorem{problem}{Problem}[section]
\newtheorem{example}{Example}[section]
\newtheorem{remark}{Remark}
\newtheorem{corollary}{Corollary}

\begin{document}

\title{Design and Experimental Validation of Closed-Form CBF-Based Safe Control for Stewart Platform Under Multiple Constraints}


\author{Benedictus C. G. Cinun,
    Tua A. Tamba, 
    Immanuel R. Santjoko, 
    Xiaofeng Wang, 
    Michael A. Gunarso, 
    Bin Hu 
    \thanks{
    B. C. G. Cinun is with the Department of Electrical and Computer Engineering at University of Houston, Houston, TX 77004, USA. ({\tt\small bgerodacinun@uh.edu})}%
    \thanks{
    T. A. Tamba, I. R. Santjoko, and M. A. Gunarso are with the Department of Electrical Engineering, Faculty of Engineering Technology, Parahyangan Catholic University, Bandung, Indonesia ({\tt\small ttamba@unpar.ac.id, 6152001008@student.unpar.ac.id})}%
    \thanks{
    X. Wang is with Department of Electrical Engineering, University of South Carolina, Columbia, SC 29208, USA ({\tt\small wangxi@cec.sc.edu)}}%
    \thanks{
    B. Hu is with Department of Engineering Technology, Electrical and Computer Engineering, University of Houston, Houston, TX 77004, USA ({\tt\small bhu11@central.uh.edu)}}}

\markboth{Journal of \LaTeX\ Class Files,~Vol.~14, No.~8, August~2021}%
{Shell \MakeLowercase{\textit{et al.}}: A Sample Article Using IEEEtran.cls for IEEE Journals}

\IEEEpubid{0000--0000/00\$00.00~\copyright~2021 IEEE}

\maketitle

\begin{abstract}
This letter presents a closed-form solution of Control Barrier Function (CBF) framework for enforcing safety constraints on a Stewart robotic platform.
The proposed method simultaneously handles multiple position and velocity constraints through an explicit closed-form control law, 
eliminating the need to solve a Quadratic Program (QP) at every control step and enabling efficient real-time implementation.
This letter derives necessary and sufficient conditions under which the closed-form expression remains non-singular, thereby ensuring well-posedness of the CBF solution to multi-constraint problem.
The controller is validated in both simulation and hardware experiments on a custom-built Stewart platform prototype, demonstrating safety-guaranteed performance that is comparable to the QP-based formulation, while reducing computation time by more than an order of magnitude.
The results confirm that the proposed approach provides a reliable and computationally lightweight framework for real-time safe control of parallel robotic systems. The experimental videos are available on the
\href{https://nail-uh.github.io/StewartPlatformSafeControl.github.io/}{project website}\footnote{Project Website: \url{https://nail-uh.github.io/StewartPlatformSafeControl.github.io/}}.
\end{abstract}

\begin{IEEEkeywords}
Closed-form Solution, Control Barrier Function, Stewart Platform, Real-time Safe Control, Quadratic Program
\end{IEEEkeywords}

\section{Introduction}
\IEEEPARstart{S}{afety}-critical
control has become increasingly important in modern robotic systems, where state, velocity, and actuator limits must be enforced during real-time operation \cite{annaswamy2024control}.
Ensuring safety is particularly challenging when multiple constraints must hold simultaneously, as robotic systems typically exhibit nonlinear coupling, configuration-dependent inertia, and complex input–state interactions.
Control Barrier Functions (CBFs) provide a principled framework for guaranteeing constraint satisfaction through real-time modification of a nominal controller \cite{ames2019control}.
However, enforcing several CBF constraints usually requires solving a Quadratic Program (QP) at every control step, which can be computationally prohibitive for systems with fast dynamics, stringent timing requirements, or embedded system hardware.
Moreover, when multiple constraints interact, the associated CBF-QP may exhibit complex structure, potential singularities, or loss of convexity \cite{molnar2023composing, cristofaro2022safe, verginis2019closed}, making closed-form safety filters difficult to derive.

In systems with second-order dynamics, several CBF formulations have been developed to handle the relative-degree issue, including Exponential CBFs \cite{nguyen2016exponential}, High-Order CBFs (HOCBFs) \cite{xiao2021high}, energy-based CBFs \cite{kolathaya2023energy, singletary2021safety}, and reduced-order methods \cite{molnar2023safety}.
These methods enable position- and velocity-based safety constraints to be expressed in an affine form and incorporated into QP-based safety filters.
Despite these advances, solving a QP at every timestep becomes increasingly burdensome as the number of constraints grows, motivating closed-form alternatives. 
Closed-form CBF filters offer significant computational advantages and have been demonstrated for trajectory tracking \cite{cristofaro2022safe}, multi-agent systems \cite{verginis2019closed}, and nonlinear output-feedback control \cite{dini2024closed}.
However, these methods handle only a single CBF constraint at a time. 
Extending closed-form solutions to multiple simultaneous CBFs introduces new difficulties: the gradients of different constraints can become linearly dependent, causing singularities in the Karush–Kuhn–Tucker (KKT) formulation and preventing a valid analytical expression.

\IEEEpubidadjcol

To address multi-constraint scenarios, smooth composition tools such as the Log-Sum-Exp (LSE) approximation \cite{molnar2023composing, choudhary2022energy, breeden2022compositions} have been used to construct differentiable approximations of the $\min$ operator.
LSE-based CBFs have enabled the design of safe reinforcement learning \cite{wang2025multi}, neural controller shielding \cite{ferlez2020shieldnn}, input-constrained safety filters \cite{rabiee2024closed}, and multi-constraint compositions \cite{breeden2022compositions, rabiee2024composition}.
Because LSE preserves differentiability, it allows several CBFs to be merged into a single smooth constraint that can be processed within a closed-form safety filter.
Recent work has also examined multi-CBF compatibility \cite{cohen2025compatibility}, addressing how multiple constraints may conflict or reinforce one another.
These developments motivate the construction of closed-form safe controllers capable of handling multiple CBF constraints simultaneously.

The need for a multi-constraint closed-form controller becomes especially clear in robotic systems that exhibit strong nonlinear coupling and tight workspace limits.
Parallel mechanisms, in particular, impose stricter safety requirements due to their configuration-dependent dynamics, actuator interactions, and constrained feasible motion sets.
These characteristics make them an ideal testbed for evaluating whether multi-constraint closed-form CBF controllers remain computationally efficient, non-singular, and practically deployable in real time.
The Stewart platform serves as a compelling application and experimental testbed for such developments.
This six-degree-of-freedom parallel manipulator is widely used in motion simulators \cite{stewart1965platform}, marine wave compensation \cite{qiu2024modeling, chen2023dynamics}, surgical robotics \cite{kizir2019design}, and astronomical instrumentation \cite{liang2022kinematics, kazezkhan2023performance}.
Its strong coupling, strict workspace limits, and nonlinear dynamics make it an excellent testbed environment to develop and evaluate multi-constraint safety-critical controllers.

In this letter, we address these gaps by developing a unified, computationally efficient safety–critical controller for the Stewart platform.
Our contributions are threefold:
\begin{enumerate}
    \item \textbf{Closed-form multi-constraint safety filter:} We derive a closed-form control law capable of enforcing multiple position and velocity CBF constraints simultaneously and establish the necessary and sufficient conditions that guarantee the resulting expression remains non-singular and well-posed.
    \item \textbf{Real-world validation on a Stewart platform prototype:} We evaluate the proposed closed-form controller both in simulation and on a physical Stewart platform prototype, demonstrating its practical viability on a real parallel mechanism.
    \item \textbf{LSE-based multi-constraint composition:} By incorporating an LSE-based multi-constraint composition, the closed-form controller attains an adjustable inner approximation of the corresponding CBF-QP.
\end{enumerate}
Together, these contributions demonstrate a unified, experimentally validated framework for safe, real-time control of robotic systems under multiple constraints.

\vspace{5pt}

\noindent \textbf{Notation}. Let $\mathbb{R}$ denote the real numbers and $\mathbb{Z}$ the integers. Their nonnegative parts are $\mathbb{R}_{\ge 0}\!\triangleq\{r\in\mathbb{R}:r\ge 0\}$ and $\mathbb{Z}_{\ge 0}\!\triangleq\{k\in\mathbb{Z}:k\ge 0\}$. 
For an integer $n\ge 1$, $\mathbb{R}^n$ is the $n$-dimensional real vector space. 
For $x\in\mathbb{R}^n$, $\|x\|\!\triangleq\!\sqrt{x^\top x}$ is the Euclidean norm. For $A\in\mathbb{R}^{m\times n}$, the norm induced by $\|\cdot\|$ is $\|A\|\!\triangleq\!\sup_{x\neq 0}\frac{\|Ax\|}{\|x\|}$ (operator $2$-norm). The special orthogonal group is $\mathcal{SO}(n)\!\triangleq\!\{R\in\mathbb{R}^{n\times n}:R^\top R=I,\ \det R=1\}$. 
A function $\alpha:[0,a)\!\to\![0,\infty)$ is of \emph{class $\mathcal{K}$} if it is continuous, strictly increasing, and $\alpha(0)=0$.
A function $\alpha$ is of \emph{class $\mathcal{K}_\infty$} if, in addition to being a class $\mathcal{K}$, for $a=\infty$ it satisfies $\alpha(s)\!\to\!\infty$ as $s\!\to\!\infty$. 
A function $\alpha:\mathbb{R}\!\to\!\mathbb{R}$ is an \emph{extended class $\mathcal{K}$} function $\mathcal{K}_e$ if it is continuous and strictly increasing  with $\alpha(0)=0$.
The boundary and interior of a set $S\in\mathbb{R}^n$ are denoted with $\partial S$ and $\text{Int}(S)$, respectively.

\section{Preliminaries and Problem Formulation}
This section reviews some background on the Stewart platform model and the CBF concept and then formulates the problem to address.

\subsection{Stewart Platform Model}
\begin{figure}[!t]
    \centering
    \subfigure[]{
        \includegraphics[width=.7\linewidth]{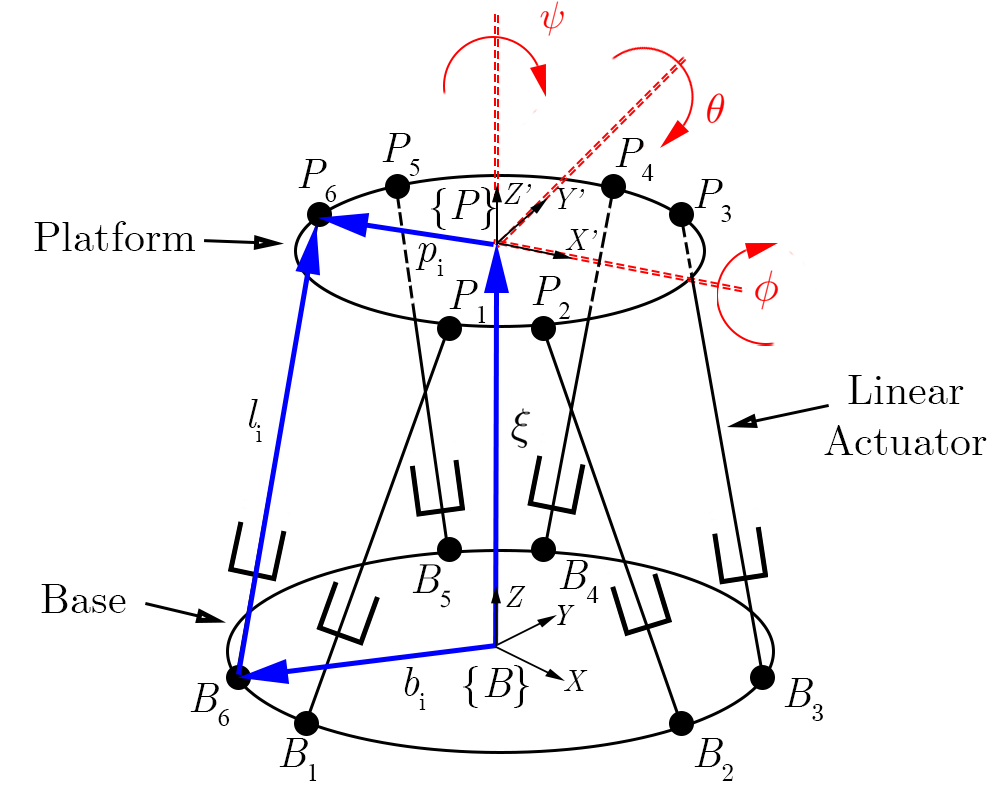}
        \label{fig:schem_kin}
    }\hspace{0.05\linewidth}
    \subfigure[]{
         \includegraphics[width=.45\linewidth]{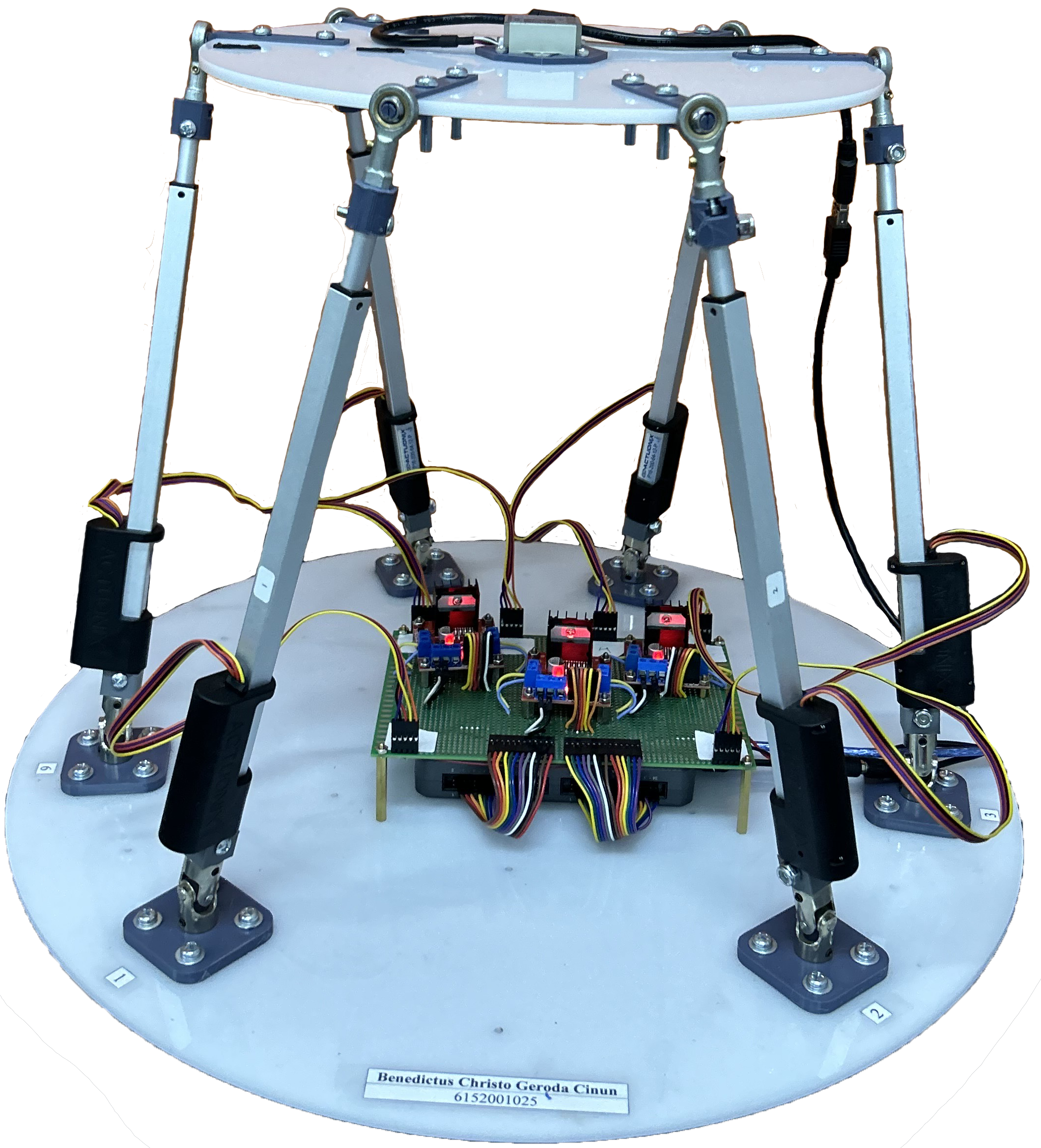}
         \label{fig:real_SP}
    }
    \caption{\subref{fig:schem_kin} The schematic of the Stewart platform, and \subref{fig:real_SP} a snapshot of its developed prototype \cite{cinun2025endtoenddesignvalidationlowcost}.}
    \label{fig:schem_and_real}
\end{figure}

The schematic of the Stewart robotic platform considered in this letter \cite{cinun2025endtoenddesignvalidationlowcost} is shown in Fig.~\ref{fig:schem_kin}.
It consists of a fixed base with a frame $\{B\}$, a moving platform with a frame $\{P\}$, and six linear actuators that connect the attachment points $B_i$ on the base to the points $P_i$ on the platform, $i=1,\dots,6$. 
The position vectors of the attachments relative to their frame origins are $b_i$ and $p_i$, respectively. 
The pose of the platform is $q = [\,\xi^\top,\ \eta^\top\,]^\top \in \mathbb{R}^6$, where $\xi=[X',Y',Z']^\top$ and $\eta=[\phi,\theta,\psi]^\top$ denote position and orientation, respectively. 
Let $R(\eta)\in \mathcal{SO}(3)$ 
be the rotation from $\{P\}$ to $\{B\}$ \cite{guo2006dynamic}. 
The $i$-th actuator vector and length are $l_i = \xi + R(\eta)\,p_i - b_i.$
The platform dynamic model is given as follows \cite{guo2006dynamic,csumnu2017simulation,bingul2012dynamic, cinun2025endtoenddesignvalidationlowcost}:
%
\begin{equation}
\mathrm{M}(q)\ddot q+\mathrm{C}(q,\dot q)\dot q+\mathrm{G}(q)=\mathrm{H}(q)\mathrm{F},
\label{eqn:stewart_dyn_equation}
 \end{equation}
 where $\mathrm{M}(q)$ is the inertia matrix, $\mathrm{C}(q)$ is the Coriolis/centrifugal terms, $\mathrm{G}(q)$ is the gravity vector, $\mathrm{H}(q)=\mathrm{J}(q)^{-\mathrm{T}}$ is the inverse-transposed Jacobian matrix, and $\mathrm{F}$ is the actuator forces.
Using $x=[\,q^\top,\ \dot q^\top\,]^\top\in\mathbb{R}^{12}$ as a vector of state variables, ~\eqref{eqn:stewart_dyn_equation} may equivalently be written in the following control–affine form~\cite{COHEN2024100947}
 \begin{align*}
    \underbrace{
        \begin{bmatrix}
            \dot{q}\\
            \ddot{q}
        \end{bmatrix}}_{\dot{x}}
        = 
        \underbrace{
        \begin{bmatrix}
            \dot{q} \\
            -\mathrm{M}(q)^{-1}(\mathrm{C}(q,\dot{q})\dot{q}+\mathrm{G}(q))
        \end{bmatrix}}_{f(x)}
        +
        \underbrace{
        \begin{bmatrix}
            0 \\
            \mathrm{M}(q)^{-1}\mathrm{H}(q)
        \end{bmatrix}}_{g(x)}\mathrm{F}.
    \end{align*}
This letter addresses the safe tracking problem of the Stewart platform model in \eqref{eqn:stewart_dyn_equation}, where a real-time controller is developed to handle compatibility with safety restrictions and conflicts with tracking performance. 
The proposed approach is validated in the platform prototype shown in Fig.~\ref{fig:real_SP}.  

\subsection{Control Barrier Functions for Safety}
This subsection provides the CBF background that underpins our closed-form safety filter and the subsequent compatibility analysis.
To begin with, the safety property defined in the state $x=[\,q^\top,\ \dot q^\top\,]^\top\in\mathbb{R}^{2n}$ with $q\in\mathbb{R}^n$ is specified as the intersection of superlevel sets of continuously differentiable functions $\{h_j\}_{j=1}^N$ of the following form.
\begin{equation}
\label{eqn:safe_set}
    \begin{aligned}
    S &= \bigcap_{j=1}^N \{ x\in D : h_j(x)\ge 0 \}\\
    \partial S &= \bigcup_{j}\{x: h_j(x)=0\}\\
    \mathrm{Int}(S) &= \bigcap_{j}\{x: h_j(x)>0\}
    \end{aligned}
\end{equation}
where $D\subset\mathbb{R}^n$ is the bounded domain set of $x$.
In this letter, the constraints are placed \emph{only} on the configuration and velocity states, i.e., each $h_j$ depends on $x=(q,\dot q)$.
The CBF framework essentially provides a means to evaluate the safety of the trajectories of dynamical systems in the set \eqref{eqn:safe_set}.
\begin{definition}[Control Barrier Function (CBF)~\cite{ames2019control,singletary2021safety}]
	A continuously differentiable function $h:D\to\mathbb{R}$ is a CBF for dynamical systems $\dot x=f(x)+g(x)u$ on $S_h=\{x:h(x)\ge 0\}$ with $S_h \subseteq D$
    if there exists $\alpha\in\mathcal{K}_e$
    such that \eqref{eqn:cbf_inequality} holds for all $x\in D$.
	\begin{equation}
		\label{eqn:cbf_inequality}
		\sup_{u}\ \big\{ L_f h(x) + L_g h(x)\,u \big\} \;\ge\; -\alpha\!\big(h(x)\big),
	\end{equation}
	with $L_f h = \nabla h\, f$, $L_g h = \nabla h\, g$, and $\nabla:=\partial/\partial x$.
\end{definition}

\textit{Constraint types and relative degree.} 
In this letter, all constraints are defined only on configuration and velocity states:
(i) \emph{position limits} $h_{p,j}(q)$ have a relative degree of two w.r.t.\ \eqref{eqn:stewart_dyn_equation} and are enforced via an \emph{energy-based} CBF that augments $h_{p,j}$ with the system's kinetic energy;
(ii) \emph{velocity limits} $h_{v,k}(\dot q)$ have a relative degree one and use the standard CBF condition \eqref{eqn:cbf_inequality}. 
This yields CBF inequalities that are affine in the actuator forces $\mathrm{F}$ and compatible with the second-order dynamics of the Stewart platform.
\begin{figure*}[t]
    \centering
    \includegraphics[width=1\linewidth]{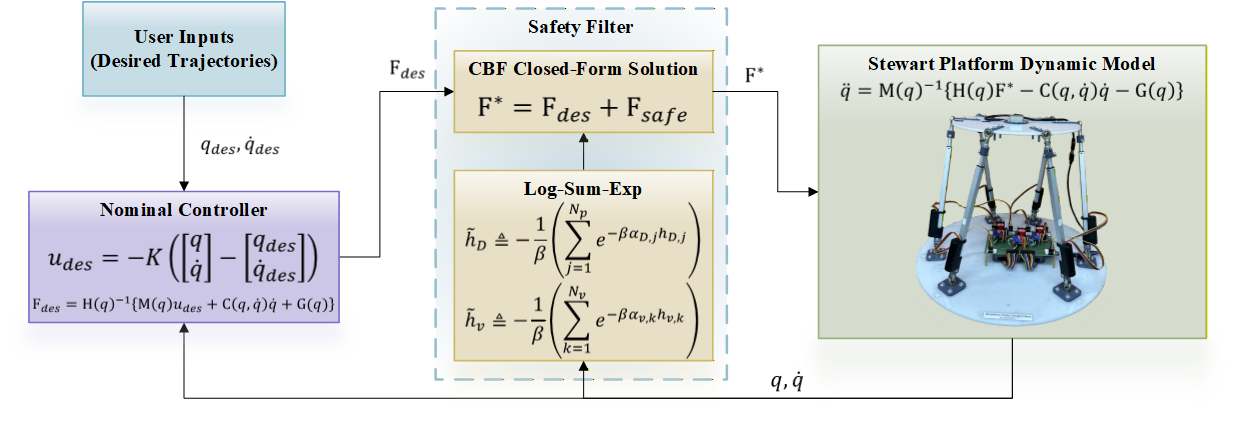}
    \caption{Architecture of the proposed Stewart platform controller. 
    The user-provided trajectory is processed by the nominal controller to generate $\text{F}_{des}$, which is filtered by the CBF-based safety module to produce the safe input $\text{F}^*$, applied to the plant with state feedback.} 
    \label{fig:cbf-architecture}
\end{figure*}


\subsection{Problem Formulation}
\label{sec:problem}

%
This letter addresses the safe tracking problem for a Stewart platform subject to safety constraints on $(q,\dot q)$. The multiple CBFs enforcing position and velocity limits, along with the corresponding problem formulation, are introduced below.

\textit{Position-limit CBF (energy-based)}:
Let $q_{\max} =[q_{\max, 1},\ldots, q_{\max, n}]^\top \in \mathbb{R}^n$ denote the component-wise upper
bounds on the configuration $q = [q_1,\dots,q_n]^\top$. For each coordinate $j$, we define the scalar
position CBF
$
    h_{p,j}(q)
    =
    q_{\max,j} - q_j.
$
This induces the box-type position safe set
\begin{equation*}
    S_p
    \,=\,
    \bigl\{ q \in \mathcal{Q} : h_{p,j}(q) \ge 0, \ \forall j \bigr\}.
\end{equation*}

Additionally, for the second-order dynamics in \eqref{eqn:stewart_dyn_equation},
we adopt the energy-based CBF proposed in~\cite{singletary2021safety}. For each
coordinate $j$, we define
\begin{equation}
    h_{D,j}(q,\dot{q})
    \,=\,
    -\tfrac{1}{2}\,\dot{q}^\top \mathrm{M}(q)\, \dot{q}
    \;+\; \alpha_e\, h_{p,j}(q),
    \label{eqn:energy_constraint}
\end{equation}
with the corresponding safe set
\begin{equation}
    S_D
    \,=\,
    \big\{ (q,\dot{q}) \in \mathcal{Q}\times\mathbb{R}^n :
           h_{D,j}(q,\dot{q}) \ge 0, \ \forall j \big\},
\end{equation}
in which $\mathcal{Q}$ is the configuration space and the parameter $\alpha_e > 0$
adjusts the conservativeness of the considered CBFs.

\textit{Velocity-limit CBF (standard)}:
Let $\dot{q}_{\max} = [\dot{q}_{\max, 1},\ldots, \dot{q}_{\max, n}]^\top \in \mathbb{R}^n$ denote the component-wise upper bounds on the velocity. For each coordinate $k$, we
define the scalar velocity CBF
\begin{equation}
\label{eq:hv}
    h_{v,k}(\dot{q})
    \,=\,
    \dot{q}_{\max,k} - \dot{q}_k.
\end{equation}
Stacking these component-wise constraints, we obtain the velocity safe set
\begin{equation*}
    S_v
    \,=\,
    \big\{ \dot{q} \in \mathbb{R}^n : h_{v,k}(\dot{q}) \ge 0, \ \forall k \big\}.
\end{equation*}
Each $h_{v,k}$ in \eqref{eq:hv} has relative degree one, so the CBF
condition in \eqref{eqn:cbf_inequality} can be evaluated directly for
each constraint \footnote{Lower-bound CBFs are obtained analogously as 
$h^{\mathrm{lo}}_{p,j}(q)\triangleq q_j-q_{\min,j}$ and 
$h^{\mathrm{lo}}_{v,k}(\dot q)\triangleq \dot q_k-\dot q_{\min,k}$. 
They are simply added as extra scalar constraints (or folded into the LSE aggregates), so all derivations and closed-form results remain unchanged. 
For notational convenience, we present only upper-bound limits in the main text.}.

\paragraph*{Safety Filter as a QP}
Given system dynamics \eqref{eqn:stewart_dyn_equation} with the CBFs $\{h_{D,j}\}$ and $\{h_{v,k}\}$ in \eqref{eqn:energy_constraint} and \eqref{eq:hv}, respectively, the safe actuator forces can be computed using the QP below.
\begin{equation}
\label{eqn:qp_cbf}
\begin{aligned}
\mathrm{F}^\star 
&= \arg\min_{\mathrm{F} \in \mathbb{R}^m} \ \big\| \mathrm{F} - \mathrm{F}_{\mathrm{des}} \big\|^2 \\
\text{s.t.}\quad 
&\qquad \dot{h}_{D,j}\ge -\,\alpha_D\big(h_{D,j}(q,\dot{q})\big), 
   \qquad \forall j, \\[0.4em]
&\qquad \dot{h}_{v,k}\ge -\,\alpha_v\big(h_{v,k}(\dot{q})\big), 
   \qquad \forall k,
\end{aligned}
\end{equation}
in which $\alpha_D,\,\alpha_v\in\mathcal{K}_e$, and $(\cdot)_k$ denotes the $k$-th component. 
The constraint functions in \eqref{eqn:qp_cbf} are defined by
\begin{equation}
    \begin{aligned}
        \dot{h}_{D,j}&=-\,\dot{q}^\top \mathrm{H}\,\mathrm{F} 
   + \mathrm{G}^\top \dot{q} 
   + \alpha_e\, \dot{h}_{p,j} \\
        \dot{h}_{v,k}&=\,\big(\mathrm{M}^{-1}\mathrm{H}\big)_{k}\,\mathrm{F} 
   + \big(\mathrm{M}^{-1}\mathrm{C}\dot{q}\big)_{k}
   + \big(\mathrm{M}^{-1}\mathrm{G}\big)_{k},
    \end{aligned}
\end{equation}
while the nominal actuator force $\mathrm{F}_{\mathrm{des}}$ in the objective function of \eqref{eqn:qp_cbf} is defined based on a nominal control input $u_{\mathrm{des}}$ as \cite{cinun2025endtoenddesignvalidationlowcost}:
\begin{equation}
	\label{eqn:F_des}
	\mathrm{F}_{\mathrm{des}} = \mathrm{H}^{-1}\!\left( \mathrm{M}\,u_{\mathrm{des}} + \mathrm{C}\,\dot q + \mathrm{G} \right).
\end{equation}
Note that \eqref{eqn:F_des} defines the unconstrained actuator forces before safety correction is considered through the CBF-QP.

\paragraph*{Nominal tracking input}
This letter assumes a nominal control input $u_{\mathrm{des}}$ for \eqref{eqn:stewart_dyn_equation} based on full-state feedback linearization  with a LQR tracking gain matrix $K$. For a desired state $[\,q_{\mathrm{des}},\dot q_{\mathrm{des}}\,]^\top$,
\begin{equation*}
	u_{\mathrm{des}} \,=\, -\,K
	\Big(
	\begin{bmatrix} q \\ \dot q \end{bmatrix}
	-
	\begin{bmatrix} q_{\mathrm{des}} \\ \dot q_{\mathrm{des}} \end{bmatrix}
	\Big),
\end{equation*}
and the corresponding nominal actuator forces follows the inverse dynamics mapping in \eqref{eqn:F_des}.
We are now ready to formally state the considered problem and goal of this letter.

\begin{problem}
	\label{prob:objective}
	Given the Stewart platform dynamics in \eqref{eqn:stewart_dyn_equation} with state $x=[q^\top,\dot q^\top]^\top$, the energy-based position CBFs $h_{D,j}(q,\dot q)$, the standard velocity CBFs $h_{v,k}(\dot q)$, and the safety-filter QP \eqref{eqn:qp_cbf} with nominal force input $\mathrm{F}_{\mathrm{des}}$, the goal of this letter is as follows.
	
	\textbf{Goal:} Derive an explicit algebraic mapping
	\[
	\Phi:\ (x,\mathrm{F}_{\mathrm{des}})\ \mapsto\ \mathrm{F}_{\mathrm{cf}}
	\]
	for the explicit/closed-form actuator forces $\mathrm{F}_{\mathrm{cf}}$ such that:
	\begin{enumerate}
		\item \textit{Inner-approximation with tunable gap:} If \eqref{eqn:qp_cbf} is feasible with feasible set $\mathcal{F}(x)$, then $\mathrm{F}_{\mathrm{cf}}$ satisfies all CBF constraints and
		\[
		\mathrm{F}_{\mathrm{cf}} \;=\; \arg\min_{\mathrm{F}\in \mathcal{F}_{\mathrm{in}}(x;\,\theta)} \ \|\mathrm{F}-\mathrm{F}_{\mathrm{des}}\|, \,
		\mathcal{F}_{\mathrm{in}}(x;\,\theta)\ \subseteq\ \mathcal{F}(x),
		\]
		where $\mathcal{F}_{\mathrm{in}}(x;\theta)$ is a parameterized inner approximation governed by tunable design parameters $\theta$ (e.g., safety margins and barrier gains such as $\alpha_e$, $\alpha_D$, $\alpha_v$). Adjusting $\theta$ trades conservativeness for tracking performance, thereby tuning the performance gap between $\mathrm{F}_{\mathrm{cf}}$ and the QP optimizer.
		\item \textit{Safety:} With input $\mathrm{F}=\mathrm{F}_{\mathrm{cf}}$, the combined safe set $S_D \cap S_v$ is forward invariant.
		\item \textit{Real-time implementability:} $\Phi$ is closed-form (no online numerical optimization) and suitable for embedded execution on the Stewart platform hardware.
	\end{enumerate}
\end{problem}
\section{Main Results}

This section presents the main results. First, Lemma~\ref{lemma} derives a closed-form, safety-preserving solution equivalent to the CBF–QP in \eqref{eqn:qp_cbf} for the single-constraint case (one position and one velocity constraint, i.e., $j=1$ and $k=1$). Building on this construction, Theorem~\ref{theorem} extends the closed-form filter to multiple simultaneous position and velocity constraints and establishes forward invariance of $S_D \cap S_v$ with minimal-norm deviation from the nominal input. The derived closed-form solutions provide a valid and tunable \emph{inner approximation} of the CBF-QP provided that the necessary and sufficient non-singularity conditions stated in Corollary~\ref{corollary} are satisfied.

\subsection{Single-Constraint Case}
Consider the Stewart platform dynamics in \eqref{eqn:stewart_dyn_equation} and the QP constraints in \eqref{eqn:qp_cbf} with one position CBF $h_{D,j}$ and one velocity CBF $h_{v,k}$.
Let $a_p(q,\dot q) \in \mathbb{R}^m$ and $a_{v,k}(q) \in \mathbb{R}^m$ be as follows.
\begin{equation}
\label{eq:ap-av1}
a_p(q,\dot q) \triangleq \mathrm{H}(q)^\top \dot q, \qquad 
a_{v,k}(q) \triangleq \big(\mathrm{M}(q)^{-1}\mathrm{H}(q)\big)^\top e_k
\end{equation}
where $e_k$ is the $k$-th standard basis vector such that $a_{v,k}$ is the transpose of the $k$-th row of $\mathrm{M}^{-1}\mathrm{H}$. 
Furthermore, define the residuals \emph{evaluated at the nominal actuator force} $\mathrm{F}_{\mathrm{des}}$ below.
\begin{equation}
\label{eq:res}
\begin{aligned}
\Psi_{p,j} &\triangleq \dot h_{D,j}(x,\mathrm{F}_{\mathrm{des}}) + \alpha_D\!\big(h_{D,j}(x)\big)\\
\Psi_{v,k} &\triangleq \dot h_{v,k}(x,\mathrm{F}_{\mathrm{des}}) + \alpha_v\!\big(h_{v,k}(x)\big)
\end{aligned}
\end{equation}
In this case, the closed-form actuator force input $\mathrm{F}_{\mathrm{cf}}$ can be derived as stated formally in Lemma \ref{lemma}.
\begin{lemma}
	\label{lemma}
	Given the dynamics of the Stewart platform in \eqref{eqn:stewart_dyn_equation} and the QP constraints in \eqref{eqn:qp_cbf} with one position CBF $h_{D,j}$ and one velocity CBF $h_{v,k}$, the closed-form actuator force is
	of the form: $\mathrm{F}_{\mathrm{cf}}^{\mathrm{sc}} = \mathrm{F}_{\mathrm{des}} + \mathrm{F}_{\mathrm{safe}}^{\mathrm{sc}},$
	in which the minimal-norm correction term $\mathrm{F}_{\mathrm{safe}}^{\mathrm{sc}}\in\mathbb{R}^m$ is defined as follows.
	\begin{equation}
		\label{eq:Fsafe_piecewise}
		\mathrm{F}_{\mathrm{safe}}^{\mathrm{sc}} =
		\begin{cases}
			\displaystyle \frac{\Psi_{p,j}}{\|a_p\|_2^2}\, a_p, & \Psi_{p,j}<0,\ \Psi_{v,k}\ge 0,\\[8pt]
			\displaystyle \frac{\Psi_{v,k}}{\|a_{v,k}\|_2^2}\, a_{v,k}, & \Psi_{p,j}\ge 0,\ \Psi_{v,k}<0,\\[8pt]
			\displaystyle A\,\mu, & \Psi_{p,j}<0,\ \Psi_{v,k}<0,\\[6pt]
			0, & \Psi_{p,j}\ge 0,\ \Psi_{v,k}\ge 0,
		\end{cases}
	\end{equation}
	where $A \triangleq [\,a_p\ \ a_{v,k}\,]\in\mathbb{R}^{m\times 2},$ in which $a_p$ and $a_{v,k}$ are defined as in \eqref{eq:ap-av1}, $\Psi_{v,k}$ and $\Psi_{p,j}$ are defined as in \eqref{eq:res}, and
	\[
	\Gamma \triangleq A^\top A =
	\begin{bmatrix}
		\|a_p\|_2^2 & a_p^\top a_{v,k}\\
		a_{v,k}^\top a_p & \|a_{v,k}\|_2^2
	\end{bmatrix},\qquad
	\mu \triangleq \Gamma^{-1}\!\begin{bmatrix}\Psi_{p,j}\\ \Psi_{v,k}\end{bmatrix},
	\]
	provided $\Gamma$ is nonsingular (i.e., $a_p$ and $a_{v,k}$ are not collinear).
\end{lemma}

\begin{IEEEproof}
The proof is based on the KKT conditions applied to the identity-weighted QP in \eqref{eqn:qp_cbf}. 
Let $\lambda_p,\lambda_v\ge 0$ denote the multipliers for the two CBF inequalities. With $\delta \mathrm{F}\triangleq\mathrm{F}-\mathrm{F}_{\mathrm{des}}$, write the constraints as
\[
-\Psi_{p,j}+a_p^\top \delta \mathrm{F}\ \le 0,\qquad
-\Psi_{v,k}+a_{v,k}^\top \delta \mathrm{F}\ \le 0,
\]
where $a_p$ and $a_{v,k}$ are as defined in \eqref{eq:ap-av1}, while $\Psi_{p,j}$ and $\Psi_{v,k}$ are defined in \eqref{eq:res}.
The Lagrangian can be written as
\begin{equation*}
    \begin{aligned}
        \mathcal{L}(\delta \mathrm{F},\lambda_p,\lambda_v)
        =&\|\delta \mathrm{F}\|_2^2
        +\lambda_p\big(-\Psi_{p,j}+a_p^\top \delta \mathrm{F}\big) \\
        &+\lambda_v\big(-\Psi_{v,k}+a_{v,k}^\top \delta \mathrm{F}\big).
    \end{aligned}
\end{equation*}
\textbf{Stationarity:} $\nabla_{\delta \mathrm{F}}\mathcal{L}=2\,\delta \mathrm{F}+\lambda_p a_p+\lambda_v a_{v,k}=0$, hence
\begin{align}
\label{eq:stationarity}
\delta \mathrm{F}=-\tfrac{1}{2}(\lambda_p a_p+\lambda_v a_{v,k}).
\end{align}
\textbf{Complementary slackness:}
\[
\lambda_p\big(-\Psi_{p,j}+a_p^\top \delta \mathrm{F}\big)=0,\qquad
\lambda_v\big(-\Psi_{v,k}+a_{v,k}^\top \delta \mathrm{F}\big)=0.
\]
\textbf{Feasibility:} $-\Psi_{p,j}+a_p^\top \delta \mathrm{F}\le 0$, $-\Psi_{v,k}+a_{v,k}^\top \delta \mathrm{F}\le 0$.

\noindent Consider the three active-set cases.

\emph{Case 1 (no active constraint):} $\lambda_p=\lambda_v=0$ gives $\delta \mathrm{F}=0$.

\emph{Case 2 (single active constraint):}
If $\lambda_p>0$ and $\lambda_v=0$, slackness enforces $a_p^\top \delta \mathrm{F}=\Psi_{p,j}$. Using the stationarity condition in \eqref{eq:stationarity} leads to
$-\tfrac{1}{2}\lambda_p\|a_p\|_2^2=\Psi_{p,j}$, hence
$\lambda_p=-2\Psi_{p,j}/\|a_p\|_2^2$ and
\[
\delta \mathrm{F}=\frac{\Psi_{p,j}}{\|a_p\|_2^2}\,a_p.
\]
Similarly, if $\lambda_p=0$ and $\lambda_v>0$, we have
\[
\delta \mathrm{F}=\frac{\Psi_{v,k}}{\|a_{v,k}\|_2^2}\,a_{v,k}.
\]

\emph{Case 3 (two active constraints):}
If $\lambda_p>0$ and $\lambda_v>0$, the slackness condition yields
\[
A^\top \delta \mathrm{F}=\begin{bmatrix}\Psi_{p,j}\\ \Psi_{v,k}\end{bmatrix},\qquad
A\triangleq[\,a_p\ \ a_{v,k}\,]\in\mathbb{R}^{m\times 2}.
\]
Combining with \eqref{eq:stationarity} gives
$-\tfrac{1}{2}A^\top A\begin{bmatrix}\lambda_p\\ \lambda_v\end{bmatrix}
=\begin{bmatrix}\Psi_{p,j}\\ \Psi_{v,k}\end{bmatrix}$.
Let $\Gamma\triangleq A^\top A$; when $\Gamma$ is invertible (necessary and sufficient conditions provided in Corollary~\ref{corollary}),
\[
\begin{bmatrix}\lambda_p\\ \lambda_v\end{bmatrix}
=-2\,\Gamma^{-1}\begin{bmatrix}\Psi_{p,j}\\ \Psi_{v,k}\end{bmatrix},
\quad\Rightarrow\quad
\delta \mathrm{F}=A\,\Gamma^{-1}\begin{bmatrix}\Psi_{p,j}\\ \Psi_{v,k}\end{bmatrix}.
\]

Setting $\mathrm{F}_{\mathrm{cf}}^{\mathrm{sc}}=\mathrm{F}_{\mathrm{des}}+\delta \mathrm{F}$ and collecting the cases yields the piecewise expression in \eqref{eq:Fsafe_piecewise}. Each case provides the minimum-norm correction consistent with its active set, which is equivalent to the KKT solution for the identity-weighted QP.
\end{IEEEproof}

\begin{remark}
When $\Psi_{p,j}<0$ and $\Psi_{v,k}<0$, \eqref{eq:Fsafe_piecewise} yields the unique minimum-norm correction $\delta \mathrm{F}$ to $\mathrm{F}_{\mathrm{des}}$ that makes both CBF inequalities tight, i.e.:
\[
\min_{\delta \mathrm{F}} \ \|\delta \mathrm{F}\|_2 
\quad \text{s.t.} \quad
A^\top \delta \mathrm{F} = 
\begin{bmatrix}\Psi_{p,j}\\ \Psi_{v,k}\end{bmatrix}.
\]
Thus $\mathrm{F}_{\mathrm{safe}}^{\textrm{sc}} = A\,(A^\top A)^{-1}[\Psi_{p,j},\,\Psi_{v,k}]^\top$ (for invertible $A^\top A$; see Corollary~\ref{corollary}). This coincides with the identity-weighted CBF–QP restricted to these two active constraints and serves as a closed-form solution to the full CBF–QP.
\end{remark}

\subsection{Multi-Constraint Case}
Note that naively extending \eqref{eq:Fsafe_piecewise} to many position and velocity constraints by active-set enumeration leads to combinatorial (exponential) case growth. 
To avoid this, we construct a smooth inner approximation of the feasible set induced by all CBF inequalities and derive a single, analytically tractable closed-form safety filter without case explosion. 
More specifically, given the Stewart platform dynamics \eqref{eqn:stewart_dyn_equation} with energy-based position CBFs $\{h_{D,j}(q,\dot q)\}_{j=1}^{N_p}$ and velocity CBFs
$\{h_{v,k}(\dot q)\}_{k=1}^{N_v}$, our construction uses soft-min aggregates based on the LSE function defined as follows.
\begin{align}
\bar h_D &\triangleq -\tfrac{1}{\beta}\log\!\Big(\sum_{j=1}^{N_p} e^{-\beta h_{D,j}}\Big), \label{eq:theo_minD}\\
\bar h_v &\triangleq -\tfrac{1}{\beta}\log\!\Big(\sum_{k=1}^{N_v} e^{-\beta h_{v,k}}\Big) \label{eq:theo_minV}
\end{align}
where $\beta>0$. 
As in the single-constraint case, the derivation in the multi-constraint case considers control-sensitivity vectors $\bar a_p(q,\dot q)\in \mathbb{R}^m$ and $\bar a_v(q)\in \mathbb{R}^m$ of the form:
\begin{equation}
\label{eq:ap-av}
\bar a_p(q,\dot q) \triangleq \mathrm{H}(q)^\top \dot q,\qquad
\bar a_v(q) \triangleq \big(\mathrm{M}(q)^{-1}\mathrm{H}(q)\big)^\top \pi_v,
\end{equation}
as well as residuals evaluated at the nominal force $\mathrm{F}_{\textrm{des}}$ below.
\begin{equation}
\label{eq:residuals}
\begin{aligned}
\bar\Psi_p &\triangleq \dot{\bar h}_D(x,\mathrm{F}_{\textrm{des}})+\alpha_D(\bar h_D),\\
\bar\Psi_v &\triangleq \dot{\bar h}_v(x,\mathrm{F}_{\textrm{des}})+\alpha_v(\bar h_v),
\end{aligned}
\end{equation}
where $\bar h_D$ and $\bar h_v$ are defined in \eqref{eq:theo_minD} and \eqref{eq:theo_minV}, respectively.
Theorem~\ref{theorem} formally states a tractable closed-form safety filter for the Stewart platform dynamics in \eqref{eqn:stewart_dyn_equation}.

\begin{theorem}
\label{theorem}
Consider the Stewart platform dynamics in \eqref{eqn:stewart_dyn_equation} with energy-based position CBFs
$\{h_{D,j}(q,\dot q)\}_{j=1}^{N_p}$ and velocity CBFs
$\{h_{v,k}(\dot q)\}_{k=1}^{N_v}$. 
Define the LSE function of the soft-min aggregates as in \eqref{eq:theo_minD}-\eqref{eq:theo_minV}.
Let $\pi_v\triangleq [\pi_{v,1},\dots,\pi_{v,N_v}]^\top$ and 
\[
\pi_{p,j} \triangleq \frac{e^{-\beta h_{D,j}}}{\sum_{\ell=1}^{N_p} e^{-\beta h_{D,\ell}}},\quad
\pi_{v,k} \triangleq  \frac{e^{-\beta h_{v,k}}}{\sum_{\ell=1}^{N_v} e^{-\beta h_{v,\ell}}}.
\]
Define further control-sensitivity vectors $\bar a_p$ and $\bar a_v$ of the form \eqref{eq:ap-av}, as well as residuals $\bar\Psi_p$ and $\bar\Psi_v$ of the form \eqref{eq:residuals}.
Let $\bar A \triangleq [\,\bar a_p\ \ \bar a_v\,]\in\mathbb{R}^{m\times 2}$,
$\bar\Gamma \triangleq \bar A^\top \bar A$, and
\[
\bar\mu \;\triangleq\; \bar\Gamma^{-1}\!\begin{bmatrix}\bar\Psi_p\\ \bar\Psi_v\end{bmatrix}.
\]
Then the closed-form actuator force for the multi-constraint case has the same structure as that in Lemma~\ref{lemma}, i.e.:
\[
\mathrm{F}_{\mathrm{cf}}^{\mathrm{mc}} \;=\; \mathrm{F}_{\mathrm{des}} + \mathrm{F}_{\mathrm{safe}}^{\mathrm{mc}},
\]
where the correction term in this case is defined as follows.
\begin{equation}
\label{eq:cf_pvcbf_LSE_struct}
\mathrm{F}_{\mathrm{safe}}^{\mathrm{mc}} =
\begin{cases}
\dfrac{\bar\Psi_p}{\|\bar a_p\|_2^2}\,\bar a_p, & \bar\Psi_p<0,\ \bar\Psi_v\ge 0,\\[8pt]
\dfrac{\bar\Psi_v}{\|\bar a_v\|_2^2}\,\bar a_v, & \bar\Psi_p\ge 0,\ \bar\Psi_v<0,\\[8pt]
\bar A\,\bar\mu, & \bar\Psi_p<0,\ \bar\Psi_v<0,\\[6pt]
0, & \bar\Psi_p\ge 0,\ \bar\Psi_v\ge 0.
\end{cases}
\end{equation}
In particular, whenever $\bar\Gamma$ is invertible (see Corollary~\ref{corollary}), the two-violation branch is well posed. 
Moreover, for $N_p=N_v=1$ (so $\pi_{p,1}=\pi_{v,1}=1$), then \eqref{eq:cf_pvcbf_LSE_struct} reduces to Lemma~\ref{lemma}.
\end{theorem}

\begin{IEEEproof}
By the LSE aggregation in \eqref{eq:theo_minD}–\eqref{eq:theo_minV}, the input sensitivities are convex combinations of the per-constraint sensitivities, i.e.:
$L_g\bar h_D=\sum_j \pi_{p,j} L_g h_{D,j}$ and
$L_g\bar h_v=\sum_k \pi_{v,k} L_g h_{v,k}$.
Hence the aggregated CBF inequalities are affine in $\mathrm{F}$ with
control-sensitivity vectors $\bar a_p$ and $\bar a_v$ as in \eqref{eq:ap-av}.
Evaluating at the nominal input $\mathrm{F}_{\mathrm{des}}$ yields residuals $\bar\Psi_p$ and $\bar\Psi_v$ of the form \eqref{eq:residuals}; such that under
$\delta \mathrm{F}\triangleq\mathrm{F}-\mathrm{F}_{\mathrm{des}}$ they update linearly as
$\bar\Psi_p-\bar a_p^\top \delta \mathrm{F}$ and
$\bar\Psi_v-\bar a_v^\top \delta \mathrm{F}$.

Consider the identity-weighted QP with these two inequalities. Applying the
KKT conditions—stationarity and complementary slackness—exactly as in the proof
of Lemma~\ref{lemma} (with the replacements
$a_p\!\to\!\bar a_p$, $a_{v,k}\!\to\!\bar a_v$ and
$\Psi_{p,j}\!\to\!\bar\Psi_p$, $\Psi_{v,k}\!\to\!\bar\Psi_v$) yields the same
three active-set cases: no-active, single-active, and two-active. Solving them
gives the piecewise closed form in \eqref{eq:cf_pvcbf_LSE_struct}, where
$\bar A\triangleq[\,\bar a_p\ \ \bar a_v\,]$,
$\bar\Gamma\triangleq\bar A^\top \bar A$, and
$\bar\mu\triangleq\bar\Gamma^{-1}[\bar\Psi_p,\ \bar\Psi_v]^\top$ for the two-active
branch (well-posed when $\bar\Gamma$ is invertible; see
Corollary~\ref{corollary}). This completes the argument, which is identical to
Lemma~\ref{lemma} after substituting LSE-aggregated quantities.
\end{IEEEproof}

\paragraph*{Weighted LSE aggregation (per-constraint scalings).} To tune the conservativeness of individual constraints, positive
scalings $\alpha_{D,j}>0$ and $\alpha_{v,k}>0$ are introduced inside the LSE \eqref{eq:theo_minD}–\eqref{eq:theo_minV} (these scalings
are \emph{distinct} from the class-$\mathcal{K}$ maps $\alpha_D(\cdot)$ and
$\alpha_v(\cdot)$) to yield
\begin{align*}
\tilde h_D
&\triangleq -\tfrac{1}{\beta}\log\!\Big(\sum_{j=1}^{N_p} e^{-\beta\,\alpha_{D,j} h_{D,j}}\Big),\\
\tilde h_v
&\triangleq -\tfrac{1}{\beta}\log\!\Big(\sum_{k=1}^{N_v} e^{-\beta\,\alpha_{v,k} h_{v,k}}\Big).
\end{align*}
Let the normalized LSE weights be defined as follows.
\begin{align*}
\pi_{p,j}^{(\alpha)} &\triangleq 
\frac{e^{-\beta\,\alpha_{D,j} h_{D,j}}}{\sum_{\ell=1}^{N_p} e^{-\beta\,\alpha_{D,\ell} h_{D,\ell}}}, \quad
\pi_{v,k}^{(\alpha)} \triangleq 
\frac{e^{-\beta\,\alpha_{v,k} h_{v,k}}}{\sum_{\ell=1}^{N_v} e^{-\beta\,\alpha_{v,\ell} h_{v,\ell}}}.
\end{align*}
Then the input sensitivities of the aggregated barriers are
\begin{align}
\tilde a_p(q,\dot q) 
&\triangleq \Big(\sum_{j=1}^{N_p} \alpha_{D,j}\,\pi_{p,j}^{(\alpha)}\Big)\, \mathrm{H}(q)^\top \dot q \in \mathbb{R}^m, \label{eq: ta_p}\\
\tilde a_v(q) 
&\triangleq \big(\mathrm{M}(q)^{-1}\mathrm{H}(q)\big)^\top \tilde\pi_v \in \mathbb{R}^m \label{eq: ta_v}
\end{align}
with $\tilde\pi_v \triangleq \big[\alpha_{v,1}\pi_{v,1}^{(\alpha)},\dots,\alpha_{v,N_v}\pi_{v,N_v}^{(\alpha)}\big]^\top.$
Moreover, the residuals evaluated at $\mathrm{F}_{\mathrm{des}}$ are of the form
\begin{align*}
\tilde\Psi_p &\triangleq \dot{\tilde h}_D(x,\mathrm{F}_{\mathrm{des}}) + \alpha_D\!\big(\tilde h_D\big),\\
\tilde\Psi_v &\triangleq \dot{\tilde h}_v(x,\mathrm{F}_{\mathrm{des}}) + \alpha_v\!\big(\tilde h_v\big).
\end{align*}
The closed-form safety filter retains the structure of
\eqref{eq:cf_pvcbf_LSE_struct} with the replacements
$\bar a_p\!\to\!\tilde a_p$, $\bar a_v\!\to\!\tilde a_v$ and
$\bar\Psi_p\!\to\!\tilde\Psi_p$, $\bar\Psi_v\!\to\!\tilde\Psi_v$.
\begin{remark}
The additional scalings $\{\alpha_{D,j}\},\{\alpha_{v,k}\}$ bias the soft-min aggregates so that
more critical constraints contribute more to the aggregated barriers. In the context
of Theorem~\ref{theorem}, this serves two purposes tied to its conditions: 
(i) it reshapes the sensitivity vectors and residuals
($\bar a_p,\bar a_v,\bar\Psi_p,\bar\Psi_v \ \to\  \tilde a_p,\tilde a_v,\tilde\Psi_p,\tilde\Psi_v$),
which can improve the conditioning of $\bar\Gamma=A^\top A$ in the two-violation branch
(see Corollary~\ref{corollary}); and (ii) it steers which constraint becomes active,
reducing oscillations near the ties and increasing the likelihood that the active-set
solution matches the QP’s KKT pattern. The closed-form structure in
\eqref{eq:cf_pvcbf_LSE_struct} is unchanged, and only the aggregated quantities are
replaced by their weighted counterparts, yielding
$\mathrm{F}_{\mathrm{safe}}^{\mathrm{mc}}=\tilde A\,\tilde\mu$ (two-violation) or the corresponding
single-constraint projections with $(\tilde a_p,\tilde a_v,\tilde\Psi_p,\tilde\Psi_v)$.
\end{remark}

\begin{figure}[tb]
    \centering
    \includegraphics[width=1\linewidth]{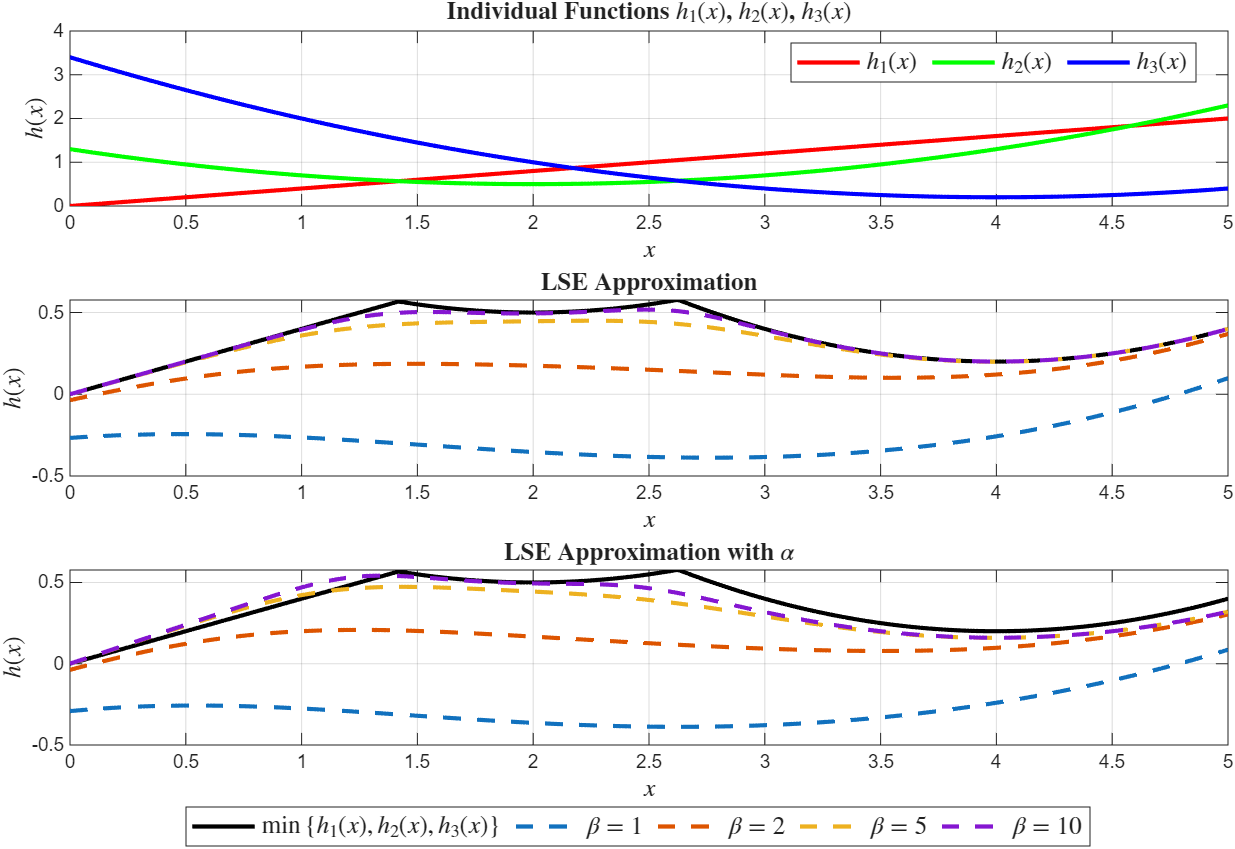}
    \caption{The approximation done with LSE produces a smooth differentiable function of a $\min$ function. The parameter $\beta$ manipulates how close the approximation is to the real $\min$ function, while $\alpha$ that is introduced to the LSE affects the scaling of the $h(x)$ functions.}
    \label{fig:LSE-example}
\end{figure}

\begin{example}
    The effect of the weighted LSE is illustrated in Fig.~\ref{fig:LSE-example} using three sample functions $h_1(x)$, $h_2(x)$ and $h_3(x)$ with assigned weights of $\alpha_1 = 1.2$, $\alpha_2 = 1$, and $\alpha_3 = 0.8$.
    While the standard LSE (middle plot) provides a uniform approximation of the minimum, the weighted version (bottom plot) shifts each function according to its assigned $\alpha$, producing a biased approximation.
    This allows specific constraints to be made more conservative or more permissive depending on their assigned weights.
\end{example}


\begin{corollary}
\label{corollary}
Assume $\mathrm{M}(q)$ and $\mathrm{H}(q)$ in \eqref{eqn:stewart_dyn_equation} are invertible. Consider the weighted LSE aggregates
with the input sensitivities of $\tilde a_p$ and $\tilde a_p$ defined in \eqref{eq: ta_p} and \eqref{eq: ta_v}, let $\tilde A=[\,\tilde a_p\ \ \tilde a_v\,]$ and $\tilde\Gamma=\tilde A^\top \tilde A$.
Then, the following are equivalent:
\begin{enumerate}
\item $\tilde\Gamma \succ 0$.
\item $\dot q \neq 0$ and $\tilde a_p \not\parallel \tilde a_v$
\; (equivalently, $a_p \not\parallel (\mathrm{M}^{-1}\mathrm{H})^\top \tilde\pi_v$).
\item $\det(\tilde\Gamma) = \|\tilde a_p\|_2^2\|\tilde a_v\|_2^2 - (\tilde a_p^\top \tilde a_v)^2 > 0$.
\end{enumerate}
These conditions specialize to Theorem~\ref{theorem} when 
	$\alpha_{D,j}\!\equiv\!1$, $\alpha_{v,k}\!\equiv\!1$
	($\bar a_p \equiv a_p$, $\bar a_v \equiv(\mathrm{M}^{-1}\mathrm{H})^\top \pi_v$),
	and to Lemma~\ref{lemma} for $N_p\!=\!N_v\!=\!1$
	($\tilde a_p \equiv a_p$, $\tilde a_v \equiv a_{v,k}$).
\end{corollary}

\begin{IEEEproof}
With $\tilde A=[\,\tilde a_p\ \ \tilde a_v\,]$ and $\tilde\Gamma=\tilde A^\top \tilde A$, recall the
construction of the weighted LSE sensitivities:
$\tilde a_p \triangleq \bar\alpha_p\, a_p,$
$\bar\alpha_p>0,$
$a_p \triangleq \mathrm{H}(q)^\top \dot q,$ and
$\tilde a_v \triangleq \big(\mathrm{M}(q)^{-1}\mathrm{H}(q)\big)^\top \tilde\pi_v,$ with
$\tilde\pi_v \succ 0$.
When then have the following.

\emph{(1) $\tilde\Gamma \succ 0$ $\Leftrightarrow$ column independence.}
For any $z\in\mathbb{R}^2$, $z^\top \tilde\Gamma z=\|\,\tilde A z\,\|_2^2\ge 0$; hence $\tilde\Gamma$ is a Gram matrix. 
It is positive definite iff $\ker(\tilde A)=\{0\}$, i.e., iff the two columns $\tilde a_p$ and $\tilde a_v$ are nonzero and not collinear. 
Since $\mathrm{M}(q)$ and $\mathrm{H}(q)$ are invertible, $(\mathrm{M}^{-1}\mathrm{H})^\top$ has trivial nullspace; with $\tilde\pi_v\neq 0$ (strictly positive), $\tilde a_v\neq 0$. 
Also $\tilde a_p=0$ iff $\dot q=0$. 
Thus $\tilde\Gamma \succ 0$ iff $\dot q\neq 0$ and $\tilde a_p\not\parallel \tilde a_v$, proving (1)$\Leftrightarrow$(2).

\emph{(2) $\Leftrightarrow$ (3).}
For $(2\times2)$ Gram matrix with columns $u$ and $v$, then 
$\det([u\ v]^\top [u\ v])=\|u\|_2^2\|v\|_2^2-(u^\top v)^2$. 
By Cauchy–Schwarz inequality, this determinant is $>0$ iff $u$ and $v$ are not collinear (and both nonzero). 
Evaluation with $u=\tilde a_p$, $v=\tilde a_v$ thus gives (2)$\Leftrightarrow$(3).

The stated specializations follow by setting the LSE scalings to one (Theorem~\ref{theorem}) or taking $N_p=N_v=1$ (Lemma~\ref{lemma}).
\end{IEEEproof}


\section{Simulation and Experiment Results}


This section presents simulation and hardware results on the Stewart platform shown in Fig.~\ref{fig:real_SP} with hardware specification summarized in Table~\ref{tab:SP_specs}
(see \cite{cinun2025endtoenddesignvalidationlowcost} for details of the end-to-end design of the prototype and nominal controller).
Across representative tracking tasks with multiple simultaneous position/velocity limits, the proposed closed-form CBF filter achieves safe tracking performance comparable to the CBF–QP baseline, maintaining forward invariance of the safe set and similar tracking errors, while reducing computation significantly.
In contrast, a nominal controller without a CBF safety filter incurs constraint violations on the same tasks.

\begin{table}[b]
    \centering
    \caption{Stewart Platform Prototype Parameters \cite{cinun2025endtoenddesignvalidationlowcost}.}
    \label{tab:SP_specs}
    \begin{tabular}{l l}
    \hline
    \textbf{Component} & \textbf{Specifications} \\ \hline
    Embedded Device & NI myRIO-1900 \\ \hline
    \multirow{5}{*}{Platform and Base} 
    & Custom-cut acrylic with 5 mm thickness\\ 
    & Real base radius: 30 cm \\ 
    & Effective base radius: 20 cm \\ 
    & Real platform radius: 15 cm \\ 
    & Effective platform radius: 16 cm \\ \hline
    Motor Drivers & L298N Motor Driver\\ \hline
    \multirow{3}{*}{Actuators} 
    & Actuonix P-16P Linear Actuator \\ 
    & with potentiometer feedback \\ 
    & and 20 cm stroke \\ \hline
    Platform joint & PHS5 Rod-End Bearing\\ \hline
    \multirow{3}{*}{Base joints}
    & Universal joint \\
    & Inner hole diameter: 10 mm \\
    & Outer diameter: 16 mm \\ \hline
    IMU Sensor & Hfi-b9 ROS IMU Module \\ \hline
    \end{tabular}
\end{table}
 
\subsection{Scenario and Parameter Setup}
Both simulation and hardware trials run for $60$\,s. The platform is initialized at $q_0 =[0,\,0,\,0.4,\,0,\,0,\,0]^\top$ with
 upper bounds enforced on both position and velocity. Specifically, the position upper bounds are set as $q_{\max,X} = q_{\max,Y} = 0.1~\mathrm{m}$, $q_{\max,Z}  = 0.5~\mathrm{m}$, and the velocity upper bounds are set to $\dot{q}_{\max,X}  = \dot{q}_{\max,Y} = 2\times10^{-3}~\mathrm{m/s}$, $\dot{q}_{\max,Z} = 10\times10^{-3}~\mathrm{m/s}$.
To exercise multi-axis limits, the platform is commanded through a sequence of static waypoints $q_{\mathrm{des}} = [X_{\mathrm{des}},Y_{\mathrm{des}},Z_{\mathrm{des}},0,0,0]^\top$, where $ X_{\mathrm{des}} = 0.1m \, \forall t \in [0, 15] $ and $X_{\mathrm{des}} = 0$ otherwise;  $Y_{\mathrm{des}} = 0.1m \, \forall t \in [15, 30]$ and $Y_{\mathrm{des}} = 0$ otherwise, and 
\begin{equation}
    Z_{\mathrm{des}} =
    \begin{cases}
        0.45, & 30 \le t < 45 \\
        0.5,  & 45 \le t < 60 \\
        0.4,  & \text{otherwise.}
    \end{cases}
    \nonumber
\end{equation}

Both simulation and experiment adopt the feedback–linearization nominal controller in \eqref{eqn:F_des} as the baseline.

\subsection{Simulation Results and Discussion}
\begin{figure}[tb]
    \centering
    \includegraphics[width=1\linewidth]{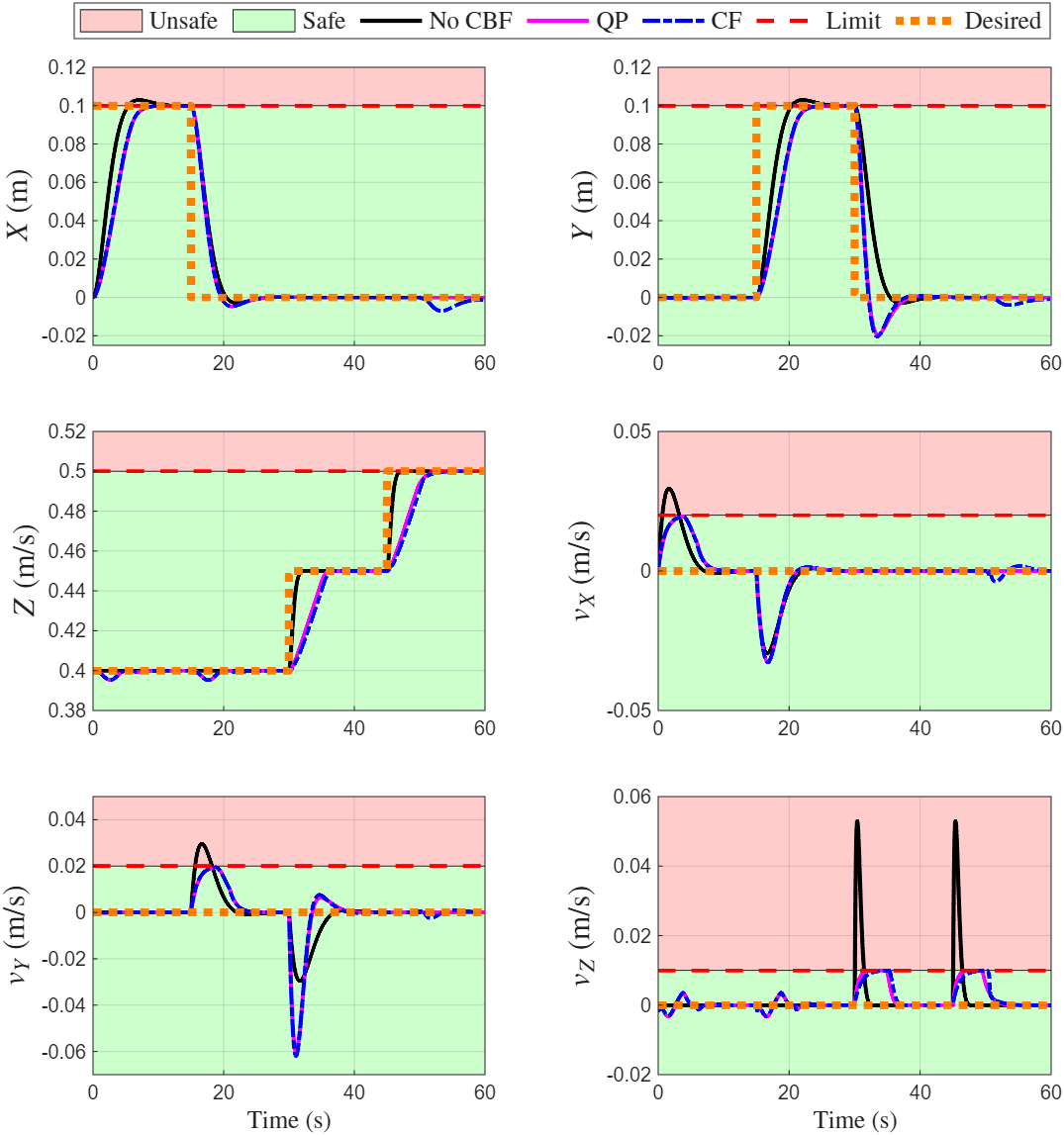}
    \caption{Simulation comparison on the Stewart platform with multiple position and velocity constraints: baseline without CBF (black solid), CBF–QP (purple solid), and closed-form filter (blue dashed). Red regions indicate the unsafe set, and green regions indicate the safe set. }
    \label{fig:sim_case2}
\end{figure}

Fig.~\ref{fig:sim_case2} shows axis velocities for waypoint tracking under three controllers: baseline (no CBF), CBF–QP, and the proposed closed-form filter. For a fair comparison, CBF parameters are identical ($\alpha_D=\alpha_e=\alpha_v=1$) with $\alpha_{v,Z}=2$. The CBF–QP (purple solid) and closed-form (blue dashed) controllers achieve comparable tracking while enforcing all position/velocity constraints throughout, demonstrating that the closed-form provides an effective, efficient alternative to online CBF–QP. By contrast, the baseline (black solid) exhibits safety violations when traversing the waypoints due to strong inter-axis coupling. The CBF-based controllers mitigate this coupling and maintain safety by adaptively trading off tracking performance.




\subsection{Experiment Results and Discussion}
\begin{figure}[tb]
 \centering
 \includegraphics[width=1\linewidth]{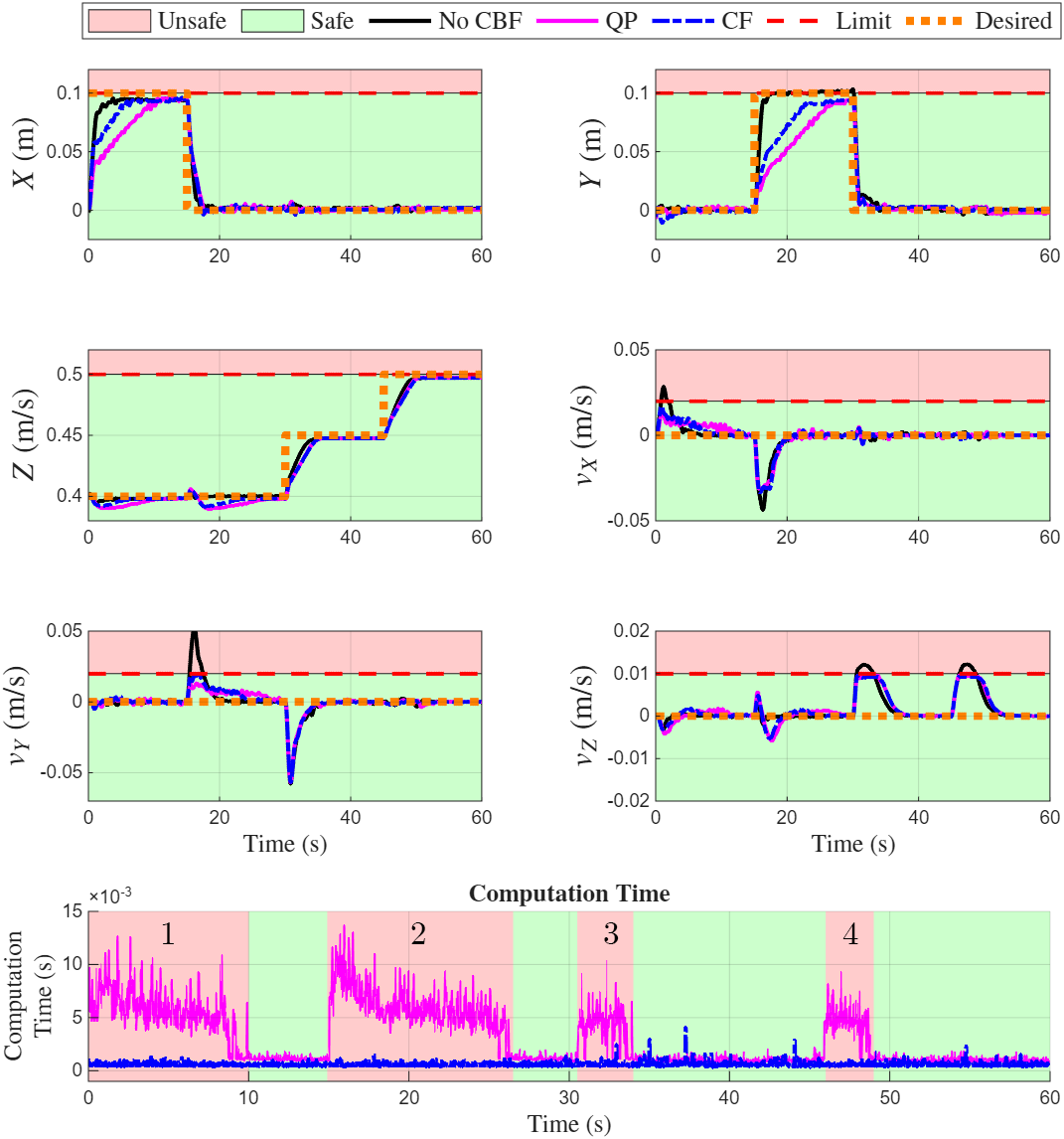}
 \caption{Experimental comparison on the Stewart platform: baseline (no CBF), CBF–QP, and closed-form CBF. Safety and tracking mirror the simulation results; an additional computation-time panel shows the closed-form is approximately $9 \times$ faster than CBF–QP in the four highlighted CBF-active regions~(when the QP is invoked).}
 \label{fig:exp_case2}
\end{figure}

In addition to the simulations, we validated the proposed closed-form CBF on the Stewart-platform hardware (Fig.~\ref{fig:schem_and_real}). Experiments used the same waypoint schedule and safety limits as in simulation. The prototype employs feedback linearization as the nominal controller, with states estimated by an Extended Kalman Filter. All CBF parameters were kept identical to simulation ($\alpha_D=\alpha_e=\alpha_v=1$) except for the $Z$-axis velocity barrier, where $\alpha_{v,Z}$ was increased from $2$ to $90$ to tighten vertical-rate regulation under hardware uncertainties.

Similar to Fig.~\ref{fig:sim_case2}, Fig.~\ref{fig:exp_case2} represents experimental axis velocities under three controllers: baseline (no CBF), CBF–QP, and the proposed closed-form filter. The $X$, $Y$, and $Z$ traces exhibit tracking behavior consistent with simulation. Both CBF-based methods trade tracking performance for safety, keeping the trajectories within the green safe region at all times, whereas the baseline violates safety during waypoint transitions. The bottom panel shows computation time: When safety constraints are active, the CBF–QP (purple solid) averages $4.98\times10^{-3}$\,s, while the closed-form (blue dashed) averages $5.83\times10^{-4}$\,s, a $\sim\!9\times$ reduction. 
Furthermore, the maximum computation time for the CBF-QP reaches $1.3720\times10^{-2}$\,s, whereas the closed-form method peaks at only $2.4270\times10^{-3}$\,s.
The details of the computation time for each violation region are shown in \ref{tab:comp_time_regions}.
These average and peak values highlighted the closed-form method’s advantage for real-time deployment\footnote{Computation times were logged in the LabVIEW control program on a laptop with 32\,GB RAM and an AMD Ryzen 8845HS processor. Deployment on lower-spec devices may yield even larger discrepancies.}
These results confirm that the proposed closed-form method delivers comparable safe tracking with substantially lower computational cost on hardware, making it attractive for real-time platforms with limited onboard resources.

\begin{table}[b]
    \centering
    \caption{Computation time statistics in the four CBF-active regions for the QP-based and closed-form CBF controllers.}
    \label{tab:comp_time_regions}
    \begin{tabular}{ccccc}
        \hline
        Region 
        & $t_{\mathrm{QP,avg}}$ (s) 
        & $t_{\mathrm{QP,max}}$ (s) 
        & $t_{\mathrm{CF,avg}}$ (s) 
        & $t_{\mathrm{CF,max}}$ (s) \\
        \hline
        1 
        & $5.51\times10^{-3}$ 
        & $1.27\times10^{-2}$ 
        & $5.94\times10^{-4}$ 
        & $1.27\times10^{-3}$ \\
        2 
        & $5.88\times10^{-3}$ 
        & $1.37\times10^{-2}$ 
        & $5.82\times10^{-4}$ 
        & $1.67\times10^{-3}$ \\
        3 
        & $4.16\times10^{-3}$ 
        & $1.04\times10^{-2}$ 
        & $5.81\times10^{-4}$ 
        & $2.43\times10^{-3}$ \\
        4 
        & $4.39\times10^{-3}$ 
        & $9.28\times10^{-3}$ 
        & $5.75\times10^{-4}$ 
        & $1.13\times10^{-3}$ \\
        \hline
    \end{tabular}
\end{table}

\section{Conclusions}
This letter presented a closed-form CBF safety filter for a 6-DoF Stewart platform that enforces multiple position and velocity constraints without per-step QP solves. The controller combines an energy-based CBF (position) with a standard CBF (velocity) and uses a modified LSE composition with an adaptive compatibility rule to manage multiple potentially conflicting constraints. Simulation and hardware experiments show safety comparable to a CBF-QP baseline with substantially lower computation—constituting, to our knowledge, the first experimental validation of a closed-form multi-constraint CBF on a parallel mechanism. These results indicate practical viability for real-time, safety-critical parallel robots. Future work will incorporate input constraints, strengthen robustness to modeling errors and disturbances, and develop principled (possibly adaptive) procedures for CBF parameter selection to balance safety and performance.

\balance
 \bibliographystyle{ieeetr}
  \bibliography{references}

\newpage

 




\vfill

\end{document}